\pdfoutput=1
\documentclass[11pt]{article}
\usepackage{acl}
\usepackage{times}
\usepackage{latexsym}
\usepackage[T1]{fontenc}
\usepackage{bbm}
\usepackage[utf8]{inputenc}
\usepackage{graphicx}
\usepackage{url}
\usepackage{amsmath}
\usepackage{xspace}
\usepackage{subcaption}
\newcommand\nd{VD1\xspace}
\newcommand\fin{VD2\xspace}
\newcommand\spangh{VD3\xspace}
\newcommand\argumentation{Arg.\xspace}

\usepackage{mathtools}

\DeclarePairedDelimiter\floor{\lfloor}{\rfloor}

\usepackage{adjustbox}
\usepackage{array}
\usepackage{xcolor}
\usepackage{color, colortbl}
\newcolumntype{R}[2]{%
    >{\adjustbox{angle=#1,lap=\width-(#2)}\bgroup}%
    l%
    <{\egroup}%
}
\newcommand*\rot{\multicolumn{1}{R{50}{1em}}}
\def\basiceval#1{\the\numexpr#1\relax}
\def\cca#1{\cellcolor{blue!#10}\ifnum #1>4\color{white!#10}\else\color{black!100}\fi{.#1}}
\def\ccb#1{\cellcolor{red!#10}\ifnum #1>4\color{white!#10}\else\color{black!100}\fi{-.#1}}
\def\ccaten#1{\cellcolor{blue!#1}\ifnum #1>40\color{white!100}\else\color{black!100}\fi{
\ifnum #1>10 .#1 \else .0#1 \fi
}}
\def\ccbten#1{\cellcolor{red!#1}\ifnum #1>40\color{white!#1}\else\color{black!100}\fi{
\ifnum #1>10 -.#1 \else -.0#1 \fi
}}
\def\ccbtentwo#1#2{\cellcolor{red!\basiceval{#2*5}}\ifnum #2>4\color{white!#1}\else\color{black!100}\fi{-#1.#2}}
\def\ccd#1{\cellcolor{blue!\basiceval{#1 * 2}}\ifnum #1>20\color{white!#1}\else\color{black!100}\fi{
\ifnum #1>10 .#1 \else .0#1 \fi
}}
\def\ccy#1{\cellcolor{yellow!\basiceval{#1 * 3}}\ifnum #1>50\color{white!#1}\else\color{black!100}\fi{
\ifnum #1>9 .#1 \else .0#1 \fi
}}

\definecolor{darkgreen}{HTML}{38761d}
\definecolor{darkyellow}{HTML}{ffa500}

\usepackage{microtype}

\title{Multitask Learning for Class-Imbalanced Discourse Classification}

\author{Alexander Spangher, Jonathan May \\ \\
  University of Southern California \\
  \texttt{\{spangher, jonmay\}@usc.edu} \\\And
  Sz-rung Shiang, Lingjia Deng \\ \\
  Bloomberg \\
  \texttt{\{sshiang, ldeng43\} }\\
  \texttt{@bloomberg.net} \\}

\begin{document}
\maketitle
\begin{abstract}
Small class-imbalanced datasets, common in many high-level semantic tasks like discourse analysis, present a particular challenge to current deep-learning architectures. In this work, we perform an extensive analysis on sentence-level classification approaches for the \textit{News Discourse} dataset, one of the largest high-level semantic discourse datasets recently published. We show that a multitask approach can improve 7\% Micro F1-score upon current state-of-the-art benchmarks, due in part to label corrections across tasks, which improve performance for underrepresented classes. 
We also offer a comparative review of additional techniques proposed to address resource-poor problems in NLP, and show that none of these approaches can improve classification accuracy in such a setting.
\end{abstract}

\section{Introduction}

Learning the discourse structure of text is an important field in NLP research, and has been shown to be helpful for diverse tasks such as: event-extraction \cite{choubey-etal-2020-discourse}, opinion-mining and sentiment analysis \cite{chenlo2014rhetorical}; natural language generation \cite{celikyilmaz2020evaluation}, text summarization \cite{lu2019attributed, isonuma2019unsupervised} and cross-document storyline identification \cite{rehm2019semantic}; and even conspiracy-theory analysis \cite{abbas2020politicizing} and misinformation detection \cite{zhou2020fake}.

However, even as recent advances in NLP allow us to achieve human-level performance in a variety of tasks, discourse-learning, a supervised learning task, faces the following challenges. (1) Discourse Learning (DL) tends to be a complex task, with tagsets focusing on abstract semantic concepts (human annotators often require training, conferencing, and still express disagreement \cite{das2017good}). (2) DL tends to be resource-poor, as annotation complexities make large-scale data collection challenging (Table \ref{tab:datasets_used}). To make matters worse, different discourse schemas often capture similar discourse intent with different labels, like recent corpora based on variations of Van Dijk's news discourse schema \cite{choubey-etal-2020-discourse, yarlott2018identifying,van2013news}.  (3) Classes tend to be very imbalanced. For example, of Penn Discourse Tree-Bank's 48 classes, the top 24 are 24.9 times more common than the bottom 24 on average \cite{prasad2008penn}.

There is no previous work establishing correspondences of discourse labels from one schema to another, but we hypothesize that a multitask approach incorporating multiple discourse datasets can address the challenges listed above. \textit{Specifically, by introducing complementary information from auxiliary discourse tasks, we can increase performance for a primary discourse task's underrepresented classes}.

We propose 
a multitask neural architecture (Section \ref{sct:methodology}) to address this hypothesis. 
We construct tasks from 6 discourse datasets and an events dataset (Section \ref{sct:datasets}), \textit{including a novel discourse dataset we introduce in this work}. Although different datasets are developed under divergent schemas with divergent goals, our framework is able to combine the work done by generations of NLP researchers and allows us not to ``waste'' their tagging work. 

Our experiments show that a multitask approach can help us improve discourse classification on a primary task, \textit{NewsDiscourse} \cite{choubey-etal-2020-discourse}, from a baseline performance of 62.8\% Micro F-1 to 67.7\%, an increase of $7\%$ (Section \ref{sct:exp_and_results}), with the biggest improvements seen in underrepresented classes.
On the contrary, simply augmenting training data
fails to improve performance.
We give insight into why this is occurring (Section \ref{sct:discussion}). In the multitask approach, the primary task's underpresented tags are correlated with tags in other datasets. However, if we only provide more data without any correlated labels, we overpredict the overrepresented tags. Meanwhile, we test many other approaches proposed to address class-imbalance, including using hierarchical labels \cite{silva2017improving}, different loss functions \cite{li2019dice}, and CRF-sequential modeling \cite{tomanek2009reducing}, and we observe similar negative results (Appendix \ref{app:neg_results}). Taken together, this analysis indicates that the signal from the labeled datasets is essential for boosting performance in class-imbalanced settings.


In summary, our core contributions are:

\begin{itemize}
  \item We show an improvement of $7\%$ above state-of-the-art on the \textit{NewsDiscourse} dataset, and we introduce a novel news discourse dataset with 67 tagged articles based on an expanded Van Dijk news discourse schema \cite{van2013news}.
  \item \textit{What worked and why}: we show that different discourse datasets in a multitask framework complement each other; correlations between labels in divergent schemas provide support for underrepresented classes in a primary task.
  \item \textit{What did not work and why:} pure training data augmentation failed to improve above baseline because they overpredicted overrepresented classes and thus hurt the overall performances.
\end{itemize}

\begin{table*}[t]
    \centering
    \begin{tabular}{|l|l|c|c|c|c|c|c|}
         \hline
         Dataset Name & Label & \#Docs & \#Sents & \#Tags & Alt. & Type & Imbal.\\
         \hline          
         \textit{NewsDiscourse} & \nd & 802 & 18,151 & 9 & No & MC & 3.01 \\
         Van Dijk \cite{yarlott2018identifying} & \fin & 50 & 1,341 & 9 & No & MC & 3.81 \\
         Van Dijk (present work) & \spangh & 67 & 2,088 & 12 & No & MC & 6.36 \\
         \hline
         Argumentation & \argumentation & 300 & 11,715 & 5 &  No & ML & 9.35 \\
         \hline
         Penn Discourse Treebank$^{**+}$ & PDTB-$t$ & 194 & 12,533 & 5 &  Yes & ML & 2.28 \\
         Rhetorical Structure Theory$^{**}$ & RST & 223 & 7,964 & 12 & Yes & ML & 2.90 \\
         KBP Events 2014/2015$^{**}$ & KBP & 677 & 24,443 & 4 & Yes & ML & 4.07 \\
         \hline
    \end{tabular}
    \caption{List of the datasets used in our work, an acronym, the size, number of tags ($k$), whether we processed it, whether there is one label per sentence (multiclass, MC) or multiple (multilabel, ML) and the class imbalance. 
    (Class imbalanced is calculated by: $\frac{\sum_{j=1}^{\floor*{k/2}} n_j}{\floor{k/2}}$ / $\frac{\sum_{j=\floor*{k/2} + 1}^{k} n_j} {\floor{k/2} + 1}$.
where $n_j$ is the number of datapoints labeled in class $j$, and classes are sorted such that $n_1 > n_2 > ... > n_k$). ** indicates that we filtered the dataset and + that we used a subset of tags.
    }
    \label{tab:datasets_used}
\end{table*}

\section{Datasets}
\label{sct:datasets}
Classical discourse tasks, based off of datasets like Penn Discourse Tree Bank (PDTB) \cite{prasad2008penn}, and Rhetorical Structure Theory Tree Bank (RST) \cite{carlson2003building}, focus on identifying clauses and classifying relations between pairs of them: such datasets give insight into temporal relations, causal-semantic relations, and the role of text in argument-supporting. Modern discourse tasks, based off datasets like the Argumentation (\argumentation) \cite{al2016news} and \textit{NewsDiscourse} (\nd), on the other hand, focus on classifying sentences: such datasets focus on the functional role each sentence is playing in a larger narrative, argument or negotiation.

We use $7$ different datasets in our multitask setup, shown in Table \ref{tab:datasets_used}. Four datasets are ``modern'' discourse datasets (containing sentence-level labels and no relational labels). Two datasets are ``classical'' discourse datasets: PDTB and RST (containing clausal relations). One event dataset, Knowledge Base Population Event Nuggets 2014/2015 (KBP), contains information on the presence of events in sentences.

\subsection{Van Dijk Schema Datasets (\nd, \fin, \spangh)}
The Van Dijk Schema, developed by Richard Van Dijk in 1988 \cite{van2013news}, was applied with no modifications in 2018 \cite{yarlott2018identifying} to a dataset, the \fin dataset, of 50 news articles sampled from the ACE corpus. The Van Dijk schema contains the following sentence-level discourse tags: \textit{Lede},
\textit{Main Event} (\textbf{M1}), \textit{Consequence} (\textbf{M2}), \textit{Circumstances} (\textbf{C1}), \textit{Previous Event} (\textbf{C2}), \textit{Historical Event} (\textbf{D1}), \textit{Expectation} \textbf{(D4)}, \textit{Evaluation} (\textbf{D3}) and \textit{Verbal Reaction}.

We introduce a novel news discourse dataset that also follows the Van Dijk Schema, the \spangh dataset. It contains 67 labeled news articles, also sampled from the ACE corpus without redundancy to \fin. An expert annotator labeled each article on the sentence-level. We made the following additions to the Van Dijk Schema: \textit{Explanation} and \textit{Secondary Event}. We introduce the \textit{Explanation} tag to help our annotator capture examples of ``Explanatory Journalism'' \cite{forde2007discovering} in the dataset, and we introduce the \textit{Secondary Event} after our annotator observed that secondary storylines sometimes present in news articles are not well-captured by the original schema's tags. To judge accuracy, the annotator additionally labeled $10$ articles that had been in \fin; the interannotator agreement was $\kappa = .69$.

The dataset of the primary task is the \nd dataset \cite{choubey-etal-2020-discourse}. This dataset contains $802$ tagged news articles and follows a modified Van Dijk schema for news discourse relation. Authors introduced the \textit{Anecdotal Event} (\textbf{D2}) tag and eliminated the \textit{Verbal Reaction} tag. Every sentence in \nd is tagged with both a discourse tag from the modified Van Dijk schema, as well another tag indicating \textit{Speech} or \textit{Not Speech}.

\subsection{Argumentation Dataset (\argumentation)}

A substantial volume of the content in news articles pertains not to factual assertions, but to analysis, opinion and explanation \cite{steele1996journalism}. Indeed, several classes in Van Dijk's schema (for example, \textit{Expectation} and \textit{Evaluation}) classify news discourse that is not factual, but opinion-focused.

Thus, we sought to include a discourse dataset that was focused on delineating different categories of opinion. The Argumentation dataset \cite{al2016news} is a sentence-level labeled dataset consisting of $300$ news editorials randomly selected from 3 news outlets.\footnote{aljazeera.com, foxnews.com and theguardian.com} The discourse tags the authors use to classify sentences are: \textit{Anecdote}, \textit{Assumption}, \textit{Common-Ground}, \textit{Statistics}, and \textit{Testimony}.\footnote{These tags share commonalities with Bales' Interactive Process Analysis categories, which delineate ways in which group members convince each other of arguments \cite{bales1950interaction, bales1970personality}, and have been used to analyze opinion content in news articles \cite{steele1996journalism}.}

\begin{figure}[t]
    \centering
    \includegraphics[width=1\linewidth]{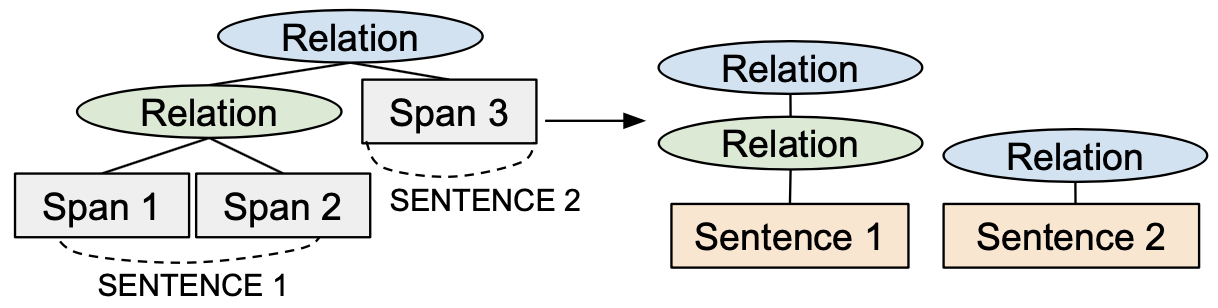}
    \caption{We processed the Penn Discourse Treebank and Rhetorical Structure Theory datasets, which are both hierarchical and relation-focused, to be sentence-level annotation tags.}
    \label{fig:pdtb_processing}
\end{figure}

\subsection{Penn Discourse Treebank (PDTB) and Rhetorical Structure Theory (RST)} 
\label{sec:pdtb_rst}
The two classical discourse datasets each identify spans of text, or clauses, and annotate how different spans relate to each other. As shown in the left-side of Figure \ref{fig:pdtb_processing}, relation annotations are between either two clauses, or between two members of the clause hierarchy. We process each dataset so that each sentence is annotated with the set of all relations occurring at least once in the sentence. We downsample each dataset so that the sentence-length distribution of articles matches the sentence-length distribution of the \nd dataset.

The full PDTB, in particular, is a high-dimensional, sparse labelset ($48$ relation classes in $2,159$ documents, $43,340$ sentences) covering a wide span of relation-types, including: \textit{Contingency}, \textit{Temporal}, \textit{Expansion} and \textit{Comparison} relations. To reduce the dimensionality and sparsity of the PDTB labelset, we only select PDTB tags pertaining to \textit{Temporal} relations, and exclude all articles that do not contain at least one such relation. This includes the tags: \textit{Temporal}, \textit{Asynchronous}, \textit{Precedence}, \textit{Synchrony}, \textit{Succession}. We filter to these tags after observing that semantic differences between certain tags in \nd depend on the temporal relation of tagged sentences: for example, the tags \textit{Previous Event} and \textit{Consequences} both describe events occurring relative to a \textit{Main Event} sentence, but a \textit{Previous Event} sentence occurs the before the \textit{Main Event} while the \textit{Consequence} occurs after. (To remove ambiguity, we henceforth refer to our filtered PDTB dataset as PDTB-$t$, for ``temporal''.)

For RST, we develop a heuristic mapping for semantically similar tags (Appendix \ref{app:datasets}) to reduce dimensionality, and then only include tags that appear in more than 800 sentences. The final set of tags that we use for RST is: \textit{Elaboration}, \textit{Joint},  \textit{Topic Change},  \textit{Attribution}, \textit{Contrast}, \textit{Explanation}, \textit{Background},
\textit{Evaluation}, \textit{Summary}, \textit{Cause}, \textit{Topic-Comment}, \textit{Temporal}.

\subsection{Knowledge Base Population (KBP) 2014/2015} 
\label{sec:kbp}

Semantic differences between certain tags in \nd depend on the presence or absence of an event, for example, the tags \textit{Previous Event} and \textit{Current Context} both provide proximal background to a \textit{Main Event} sentence, but a \textit{Previous Event} sentence contains an event while a \textit{Current Context} does not. 

We hypothesize that a dataset annotating whether events exist within a sentence can help our model differentiate these categories better. So, we collect an additional non-discourse dataset, the KBP 2014/2015 Event Nugget dataset, which annotates the trigger words for events by event-type: \textit{Actual Event}, \textit{Generic Event}, \textit{Event Mention}, and \textit{Other}. We preserve this annotation at the sentence level, similar to the PDTB and RST transformations in Section~\ref{sec:pdtb_rst} and downsample documents similarly.

\section{Methodology}
\label{sct:methodology}

\begin{figure*}[t]
    \centering
    \includegraphics[width=\linewidth]{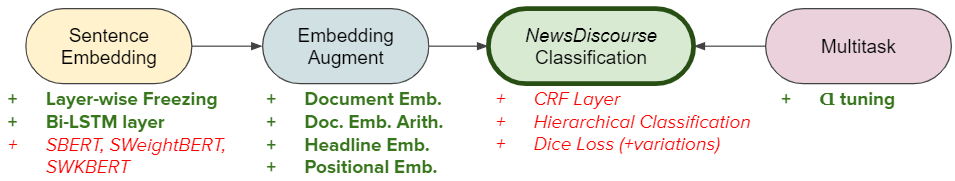}
    \caption{Overview of the experimental variations we consider, at each stage of the classification pipeline. \textcolor{darkgreen}{\textbf{Bold green}} indicates variations that had a positive effect on the classification accuracy, and \textcolor{red}{\textit{Italic red}} indicates variations that did not. Some variations are described in the Appendix.}
    \label{fig:exp_vars}
\end{figure*}

Here we describe the methods we use, along with the variations that we test, which are summarized in Figure \ref{fig:exp_vars}.

\subsection{Multitask Objective}

Our multitask setup framework can broadly be seen as a multitask feature learning (MTFL) architecture, which is common in multitask NLP applications \cite{zhang2017survey}.

\begin{figure}[t]
    \centering
    \includegraphics[width=\linewidth]{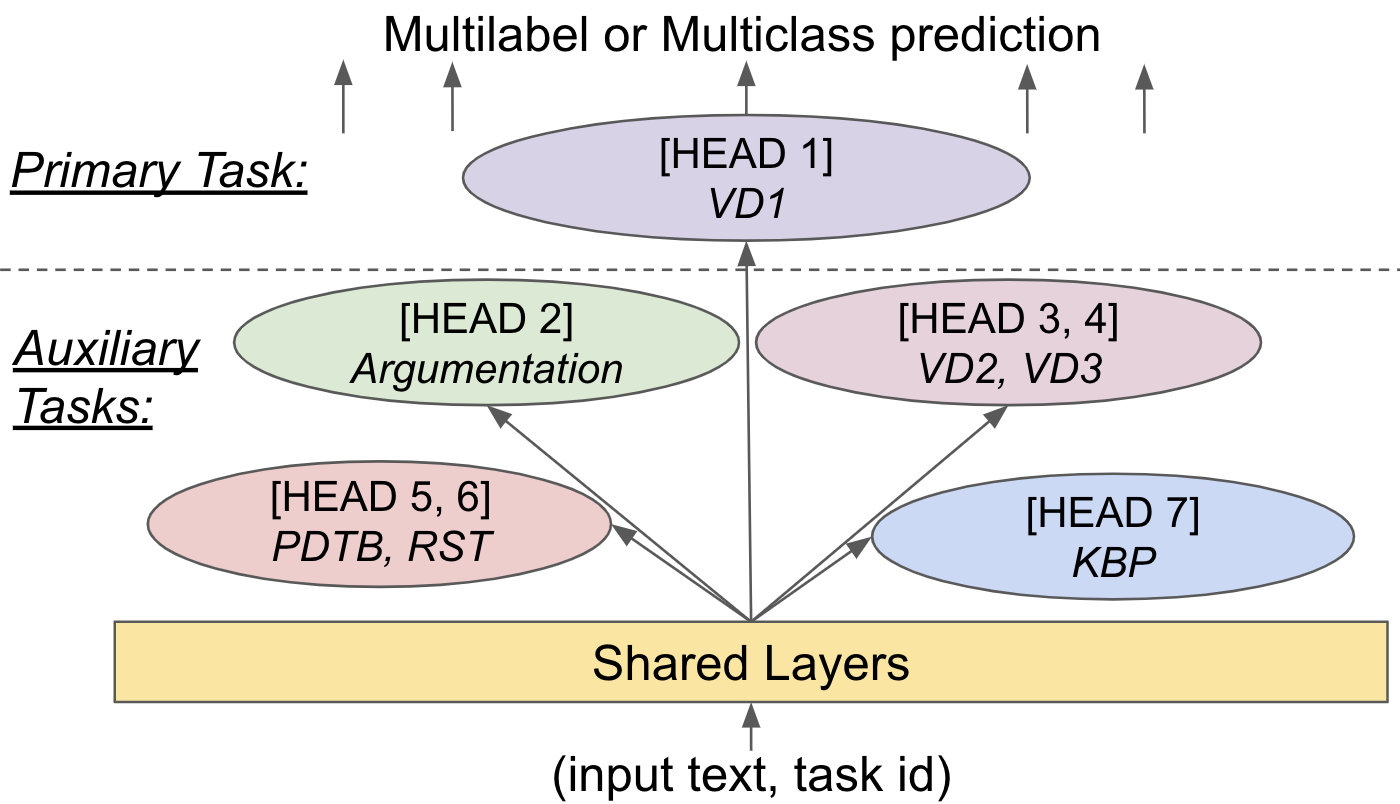}
    \caption{Our multitask setup uses 7 heads: 4 multiclass classification heads to classify \nd, \argumentation, \spangh and \fin datasets and 3 multilabel classification heads to classify PDTB, RST and KBP 2014/2015.}
    \label{fig:multitask_setup}
\end{figure}

As shown in Figure \ref{fig:multitask_setup}, we formulate a multitask approach to DL with the \nd dataset as our primary task. Our multitask architecture uses shared encoder layers and specific classification heads for each task. 

From our joined dataset $D = \{D_t\}_{t=1}^{T}$ across tasks $t=1,...,T$, where $D_t$ is of size $N_t$ and contains $\{(x_i, y_i)\}_{i=1}^{N_t}$ pairs, we randomly sample one task $t$ and one datum $(x_i, y_i)$ from that task's dataset, $D_t$. Our multitask objective is to minimize the sum of losses across tasks:

\begin{equation}
\label{eq:loss}
\min L(D, \alpha) = \min_{\theta} \sum_{t = 1}^T \sum_{i=1}^{N_t} \alpha_t L_t(D_{i}^{t})
\end{equation}

\noindent where $L_t$ is the task-specific loss with hyperparameter $\alpha = \{\alpha_t\}_{t=1}^T$, a coefficient vector summing to a constant, c, that tells us how to weight the loss from each task.

\subsection{Neural Architecture}
\label{sct:neural_architecture}
\begin{figure}[t]
    \centering
    \includegraphics[width=.9\linewidth]{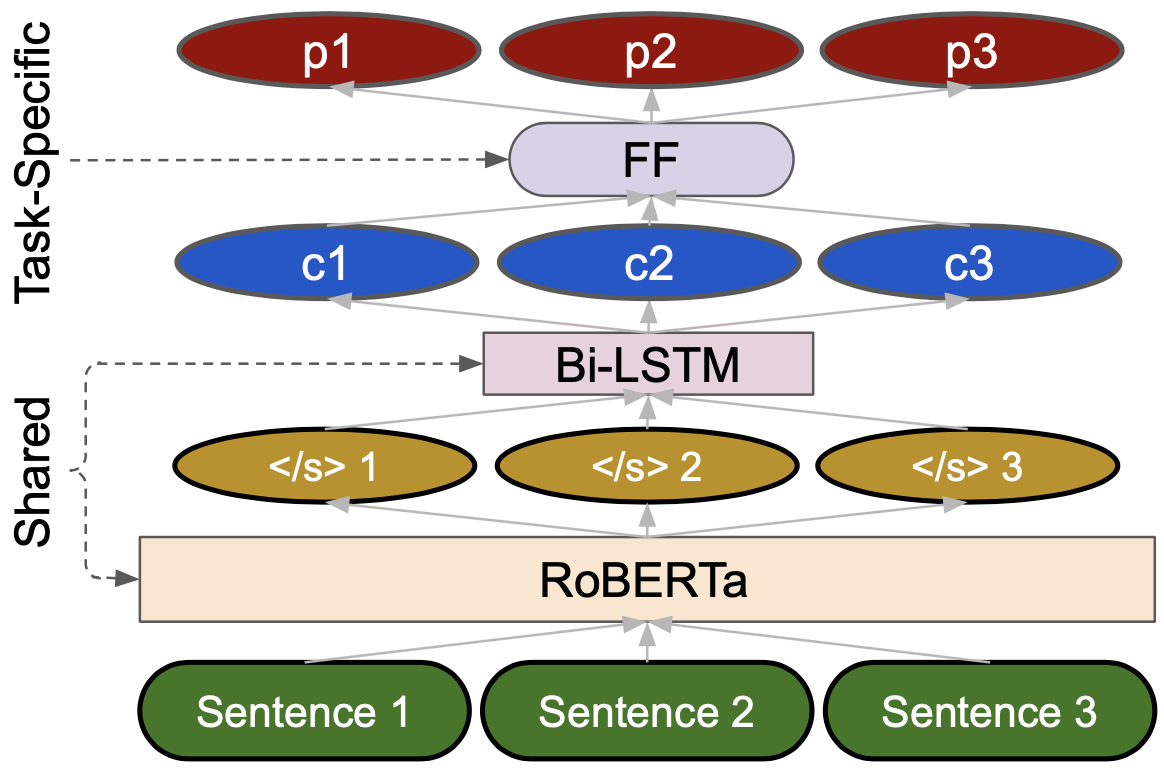}
    \caption{Sentence-Level classification model used for each prediction task. The </s> token in a RoBERTa model is used to generate sentence-level embeddings, </s>$_i$. Bi-LSTM is used to contextualize these embeddings, $c_i$. Finally, FFNN layer is used to make class predictions, $p_i$. The RoBERTa and Bi-LSTM layers are shared between tasks and the FFNN is the only task-specific layer.}
    \label{fig:nn_model}
\end{figure}

Our neural architecture, shown in Figures \ref{fig:multitask_setup} and \ref{fig:nn_model}, consists of a sentence-embedding layer with optional embedding augmentations, a classification layer for the primary task, and optional classification layers for auxiliary supervised tasks. We describe each layer in turn.

\noindent\textbf{Sentence-Embedding Modeling}
The architecture we use to model each supervised task in our multitask setup is inspired by previous work in sentence-level tagging and discourse learning \cite{choubey-etal-2020-discourse,li2019discourse}. As shown in Figure \ref{fig:nn_model}, we use a transformer model, RoBERTa-base \cite{liu2019roberta}, to generate sentence embeddings: each sentence in a document is fed sequentially into the same model, and we use the <s> token from each sentence as the sentence-level embedding. The sequence of sentence embeddings is then fed into a Bi-LSTM layer to provide contextualization. Each of these layers is shared between tasks.\footnote{Variations on our method for generating sentence embeddings are reported in Appendix \ref{app:sent_emb}}

\noindent\textbf{Embedding Augmentations}
We experiment concatenating different embeddings to our sentence-level embeddings to incorporate information on document-topic and sentence-position: headline embeddings ($H_i$) generated via the same method as sentence-embeddings; vanilla positional embeddings ($P_{i, j}$) and sinusoidal positional embeddings ($P_{i, j}^{(s)}$) as described in \citet{vaswani2017attention} but on the sentence-level rather than the word-level; document embeddings ($D_i$), and document arithmetic ($A_{i, j}$). 

To generate $D_i$ and $A_{i, j}$ for sentence $j$ of document $i$, we use self-attention on input sentence-embeddings to generate a document-level embedding, and perform the following arithmetic to isolate the topic, as done by \citet{choubey-etal-2020-discourse}:

\begin{gather}
    D_i = \text{Self-Att}(\{S_{i, j}\}_{j=1}^{N_i}) \\
    A_{i, j} = D_i * S_{i, j} \oplus D_i - S_{i, j}
\end{gather}

\noindent where $S_{i, j}$ is the sentence-embedding for sentence $j$ of document $i$, and self-attention is an operation defined by \citet{cheng2016long}. 


\noindent\textbf{Supervised Heads} Each contexualized embedding is classified using a feed-forward layer. The feed-forward layer is task-specific. The three tasks PDBT, RST, and KBP are multilabel classification tasks and the rest of the tasks are multiclass classification tasks.\footnote{Variations both of the classification task and the loss function, aimed at addressing the class imbalance inherent in the \nd dataset, are reported in Appendix~\ref{app:supervised_head_variations}.}

\subsection{Data Augmentation}
\label{sct:data_aug}

To determine whether it is the labeled information in the multitask setup that is helping us achieve higher accuracy or simply the addition of more news articles, we perform a ``data-ablation'': we test using additional data that does not contain new label information. 

We use Training Data Augmentation (TDA) to enhance single-task learning by increasing the size of the training dataset through data augmentations on the training data \cite{devries2017dataset}. 

We generate 10 augmentations for each sentence in \nd.
Our augmentation function, $g$, is a sampling-based backtranslation function, which is a common method for data augmentation \cite{edunov2018understanding}. To perform backtranslation, we use Fairseq's English to German and English to Russian models \cite{ott2019fairseq}. Inspired by \newcite{chen2020mixtext}, we generate backtranslations using random sampling with a tunable temperature parameter instead of beam search, to ensure diversity in augmented sentences.

\section{Experiments and Results}
\label{sct:exp_and_results}
In this section, we first discuss experiments using \nd as a single classification task. Then, we discuss the experiments using \nd in a multitask setting. Finally, we discuss our experiments with data augmentation.
We show explorations we did to try to maximize the performance of each method (negative results shown in Appendix \ref{app:neg_results}.)

\begin{table*}[t]
    \centering
    \begin{tabular}{|l||r|r|r|r|r|r|r|r|r||r|r|}
\hline
{} &     M1 &     M2 &     C2 &     C1 &     D1 &     D2 &     D3 &     D4 &      E &  Mac. &  Mic. \\
\hline
Support &  460 &  77 &  1149 &  284 &  406 &  174 &  1224 &  540 &  396 &   4710 &      4710 \\
\hline
\hline
ELMo    &  50.6 &  27.0 &  58.9 &  35.2 &  63.4 &  50.3 &  70.5 &  64.3 &  94.6 &  57.21 &     62.85 \\
\hline
RoBERTa &  52.1 &  9.4  &  65.1 &  27.7 &  68.1 &  51.6 &  72.4 &  65.4 &  96.0 &  56.43 &     64.97 \\
+Frozen &  51.2 &  29.3 &  64.3 &  29.8 &  72.2 &  65.8 &  73.7 &  \textbf{67.1} &  96.5 &  61.08 &     66.54 \\
+EmbAug &  54.1 &  28.0 &  64.7 &  35.9 &  71.8 &  66.3 &  72.9 &  65.9 &  96.3 &  61.76 &     66.92 \\
\hline 
\hline
TDA     &  85.3 &  52.2 &  57.1 &  29.8 &  61.1 &  44.3 &  66.1 &  58.2 &  16.4 &  56.53 &     59.22 \\
\hline 
\hline 
MT-Mac  &  54.9 &  \textbf{35.5} &  63.8 &  \textbf{35.9} &  \textbf{73.7} &  \textbf{70.7} &  \textbf{73.7} &  66.3 &  \textbf{96.7} & \textbf{63.46} &     67.51 \\
MT-Mic  &  \textbf{55.4} &  25.0 & \textbf{67.1} &  32.8 &  72.5 &  68.9 &  73.6 &  65.8 &  96.0 &  61.89 &      \textbf{67.70} \\
\hline
\end{tabular}
    \caption{\textbf{Overview:} F1-scores of individual class tags in \nd and Macro-averaged F1-score (Mac.) and Micro F1-score (Mic.). \textbf{ELMo} is the baseline used in \cite{choubey-etal-2020-discourse}. \textbf{RoBERTa+Frozen+EmbAug} is our subsequent baseline. \textbf{TDA} refers to Training Data Augmentation. MT stands for multitask: MT-Mac is a trial with $\alpha$ chosen to maximize Macro F1-score while MT-Mic is a trial with $\alpha$ chosen to maximize Micro F1-score.}
    \label{tab:positive_results}
\end{table*}

\subsection{Single Task Experiments}


\noindent\textbf{ELMo vs RoBERTa}
The first improvement we observe derives from the use of RoBERTa as a contextualized embedding layer rather than ELMo \cite{peters2018elmo}, as used in the baseline \cite{choubey-etal-2020-discourse}. We observe a 2-point F1-score improvement from 62 F1-score to 64 F1-score.

\begin{figure}[t]
    \centering
    \includegraphics[width=\linewidth]{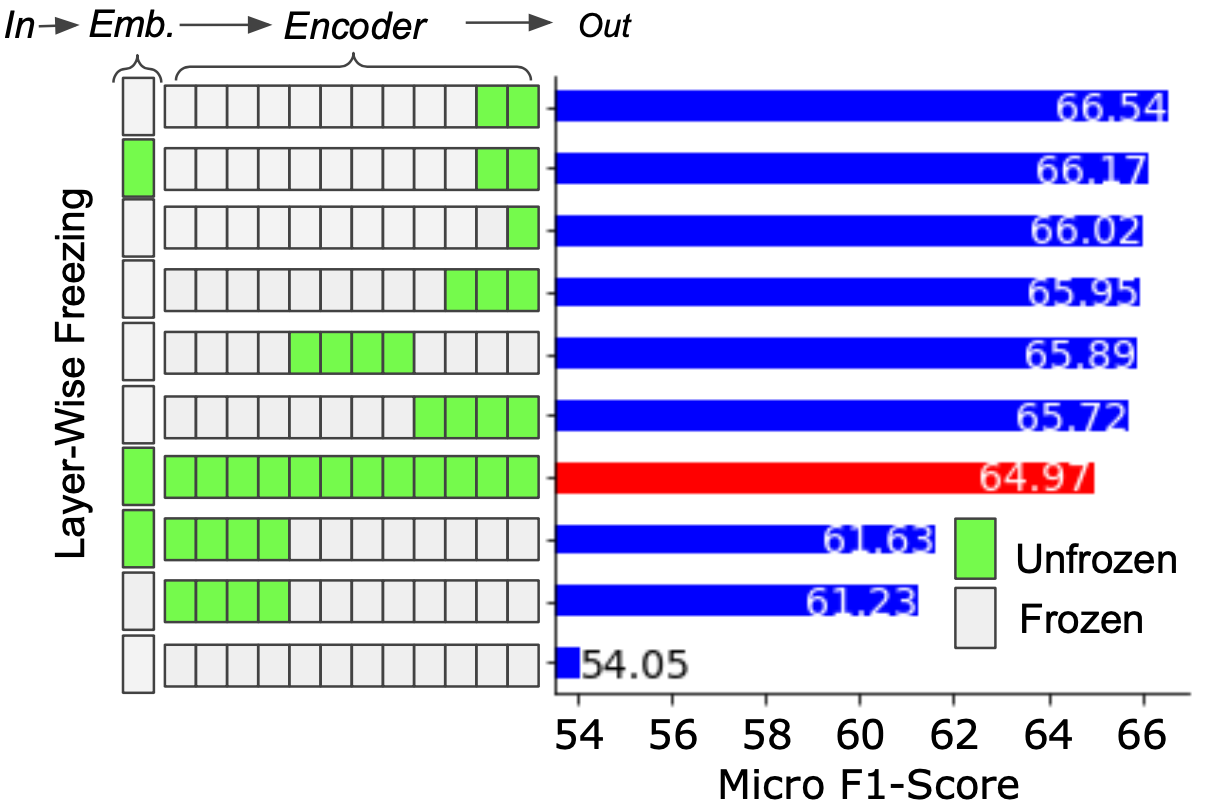}
    \caption{Here we show a sample of the different layer-wise freezing that we performed. ``Emb.'' block is the embedding lookup table for word-pieces. ``Encoder'' blocks closer to the input are visualized on the left, and blocks to the right are closer to the output. The \textcolor{red}{red} bar indicates the baseline, unfrozen RoBERTa model.}
    \label{fig:layerwise_freezing}
\end{figure}

\noindent\textbf{Layer-wise Freezing (+Frozen)}
The second improvement we observe follows from layerwise freezing for RoBERTa. We observe that unfreezing the layers closest to the output results in the greatest improvement (Figure \ref{fig:layerwise_freezing}). Layer-wise freezing was a bottleneck to all other improvements observed: prior to performing layer-wise freezing, none of the experiments we tried, including multitask, had any effect. We observe a 1.5 F1-score improvement above a RoBERTa unfrozen baseline.

\begin{table}[t]
\centering
\begin{tabular}{|l|r|}
\hline
Embedding Augmentations &         $\delta$ Micro F1 \\
\hline
$\oplus P_i^{(s)} \oplus D_i \oplus A_i \oplus H_i$ & \ccaten{38} \\
$\oplus P_i^{(s)} \oplus D_i \oplus A_i$            & \ccaten{37} \\
$\oplus P_i^{(s)} \oplus D_i \oplus H_i$            & \ccaten{35} \\
$\oplus P_i \oplus D_i \oplus A_i \oplus H_i$       & \ccaten{33} \\
\hline
\hline
$\oplus H_i$                                        & \ccaten{11} \\
$\oplus D_i$                                        & \ccaten{0} \\
$\oplus D_i \oplus A_i$                             & \ccaten{0} \\
$\oplus P_i$                                        & \ccbten{1} \\
$\oplus P_i^{(s)}$                                  & \ccbten{8} \\
\hline
\end{tabular}
\caption{\textbf{Sample} of combinations of embedding augmentation variations. Micro F1-score increase gained by adding the embedding augmentation above \textbf{+Frozen}. $P_i^{(s)}$ is sinusoidal and $P_i$ is vanilla positional embeddings. $D_i$ is document embeddings and $A_i$ is document embeddings arithmetic. $H_i$ is headline embeddings. }
\label{tbl:embedding_augmentations}
\end{table}

\noindent\textbf{Embedding Augmentations (+EmbAug)}
The third improvement we observe derives from concatenating embedding layers that contain additional document-level information. We observe a .5 F1-score improvement above a RoBERTa partially-frozen baseline with a full concatenation of Document Embeddings, Headline Embeddings, and Sinusoidal Positional Embeddings (Table \ref{tbl:embedding_augmentations}). The .5 F1-score improvement holds across different sentence embeddings variations (Appendix \ref{app:neg_results}). The embeddings appear to interact as together they present an improvement but by themselves they offer no improvement.

\subsection{Multi-Task Experiments}
\label{subsct:multitask_exp}

\begin{figure}[t]
    \centering
    \includegraphics[width=1\linewidth]{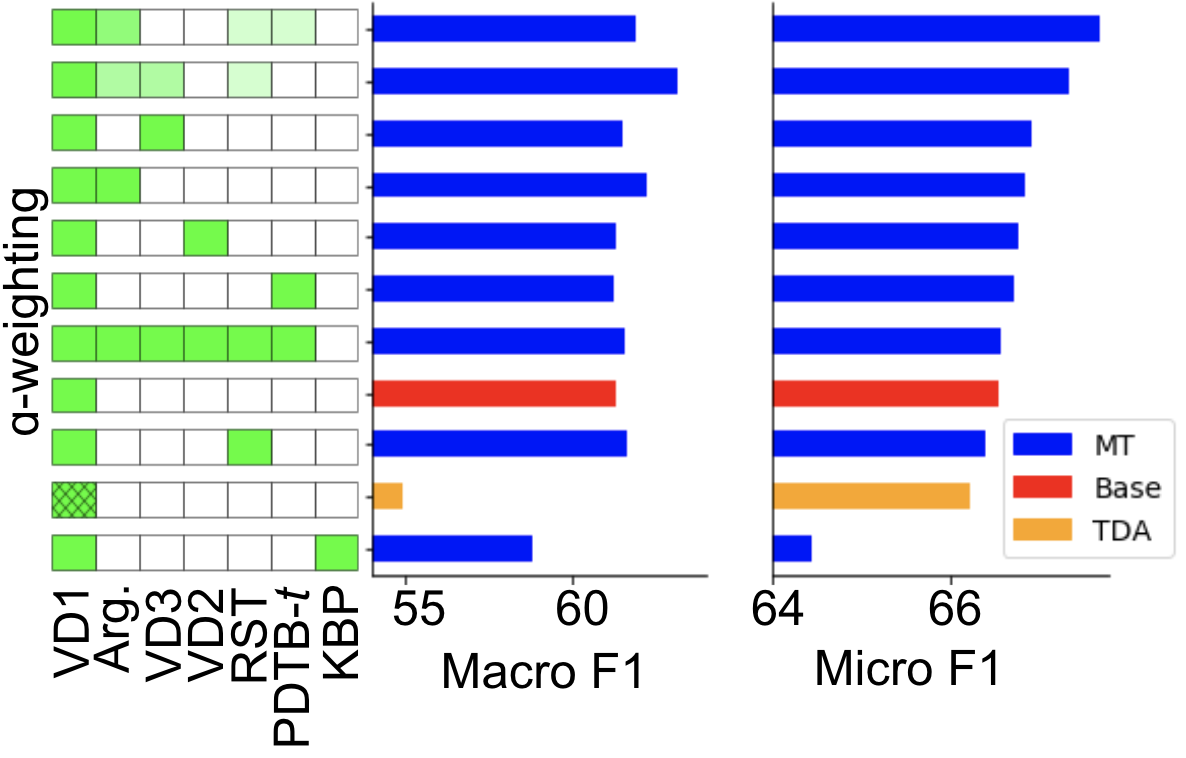}
    \caption{Loss-coefficient weightings ($\alpha$ vector) (y-axis) and Macro vs. Micro F1 Score shown for: (a) a mix of trials, (\textcolor{blue}{blue} bar, $\{$VD1$, X_1, ... X_k\}$) (b) pairwise multitask tasks (\textcolor{blue}{blue} bar, $\{$VD1$, X_1\}$), (c) baseline (\textcolor{red}{red} bar: $\{$VD1$\}$) (d) data ablation (\textcolor{darkyellow}{yellow} bar, TDA). Cells corresponding to training data sets are \textcolor{darkgreen}{green} in strength proportional to their $\alpha$ value.  Although several simple multitask settings (0-1 $\alpha$ vectors) beat the baseline, the best performing settings are a soft mix of the $6$ discourse tasks.}
    \label{fig:multitask_coefs}
\end{figure}

As shown in Table \ref{tab:positive_results}, multitask achieves the best performance.  We conduct our multitask experiment by performing a grid-search over a wide range of loss-weighting, $\alpha$ (defined in Equation \ref{eq:loss}). As can be seen, in Figure \ref{fig:multitask_coefs}, the weighting achieving the top Micro F1-score includes datasets: VD1, \argumentation, RST and PDTB-$t$, while the weighting achieving the top Macro F1-score includes datasets: VD1, \argumentation, \spangh, and RST.

We parse the effect of different loss-weighting schemes for each dataset on the output scores by running Linear Regression, where $X=\alpha$, the loss-weighting scheme, and $y=\text{F1-score}$. The Linear Regression coefficients, $\beta$, displayed in Table \ref{tbl:lin_reg}, approximate the effect of each dataset has. Note that this is just an approximation, and does not account for nonlinearity in $\alpha$ (i.e. datasets that have a negative effect within one range of $alpha$ might be less negative, or even positive within another).

\begin{figure}[t]
    \centering
    \includegraphics[width=\linewidth]{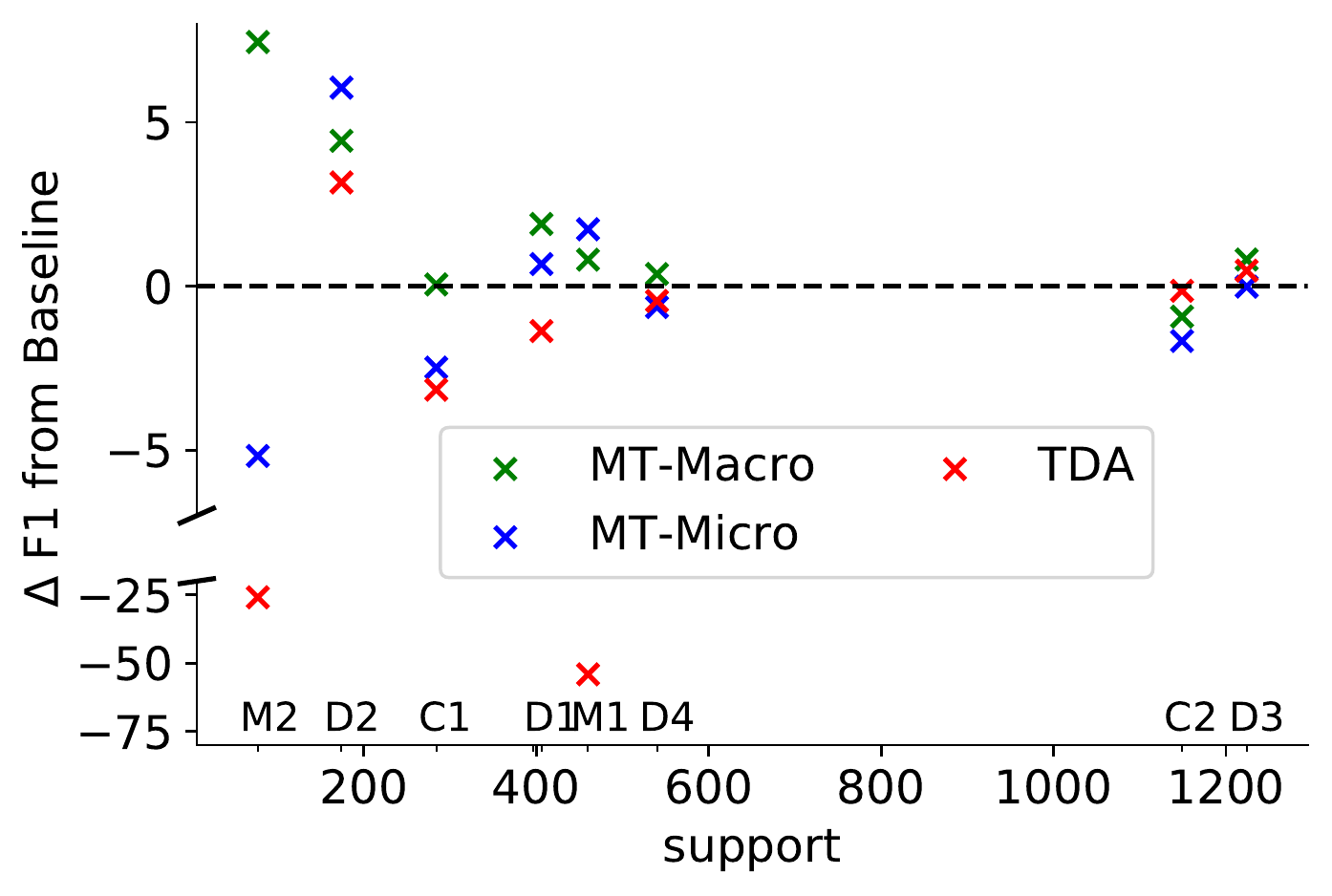}
    \caption{Comparison of class-level accuracy vs. tag for three models: MT-Micro, TDA (which underperforms baseline for lower-represented tags like M2, C1), and MT-Macro (which overperforms baseline for lower represented tags M1, M2, D1, D2). Split y-axis shown for clarity, due to TDA outliers.}
    \label{fig:scatterplot}
\end{figure}

\begin{table}[]
    \centering
    \begin{tabular}{|l|r||l|r|}
    \hline
    Dataset &      LR $\beta$ & Dataset &     LR $\beta$ \\
    \hline
    \nd (Main)           &   \ccaten{83}  & \argumentation &   \ccaten{5}  \\ 
    RST            &   \ccaten{50}  & PDTB           &  \ccbten{69}  \\
    \spangh        &   \ccaten{49}  & KBP            &  \ccbtentwo{2}{17}  \\
    \fin           &   \ccaten{21}  & Intercept      &    66.26 \\
    \hline
    \end{tabular}
    \caption{The Linear Regression coefficients ($\beta$) for each dataset to parse the effects of each dataset on the scores. We run a simple Linear Regression model, LR, on the $\alpha$ weights from all grid-search trials (a subset of which are shown in Figure \ref{fig:multitask_coefs}) to predict Micro F1-scores (i.e. LR$(\alpha)= \mbox{Mic. F1-score}$). This shows us, for example, that increasing RST's weight by $+1$ yields $.5$ F1-score improvement. Note: this is only an approximation, and dataset-weights might have a non-linear effect.}
    \label{tbl:lin_reg}
\end{table}

\subsection{Data Augmentation Experiments}
\label{sct:uda_experiments}

As shown in Table \ref{tab:positive_results} and Figure \ref{fig:multitask_coefs}, \textbf{TDA} fails to improve performance from our baseline. In the next section, we give insights into why multitask improves the performances but training data augmentation fails to do so. 

\section{Discussion}
\label{sct:discussion}

As shown in Figure \ref{fig:scatterplot}, a multitask approach to this problem significantly increases performance for classes with a lower support, while not jeopardizing the performance for classes with a higher support. This is in contrast to purely data augmentation approaches like TDA. Improving performance in low-support classes improves overall Macro F1, as expected, but, as can be seen in Table~\ref{tab:positive_results}, is beneficial to Micro F1 as well.

\begin{figure}[t]
    \centering
    \includegraphics[width=\linewidth]{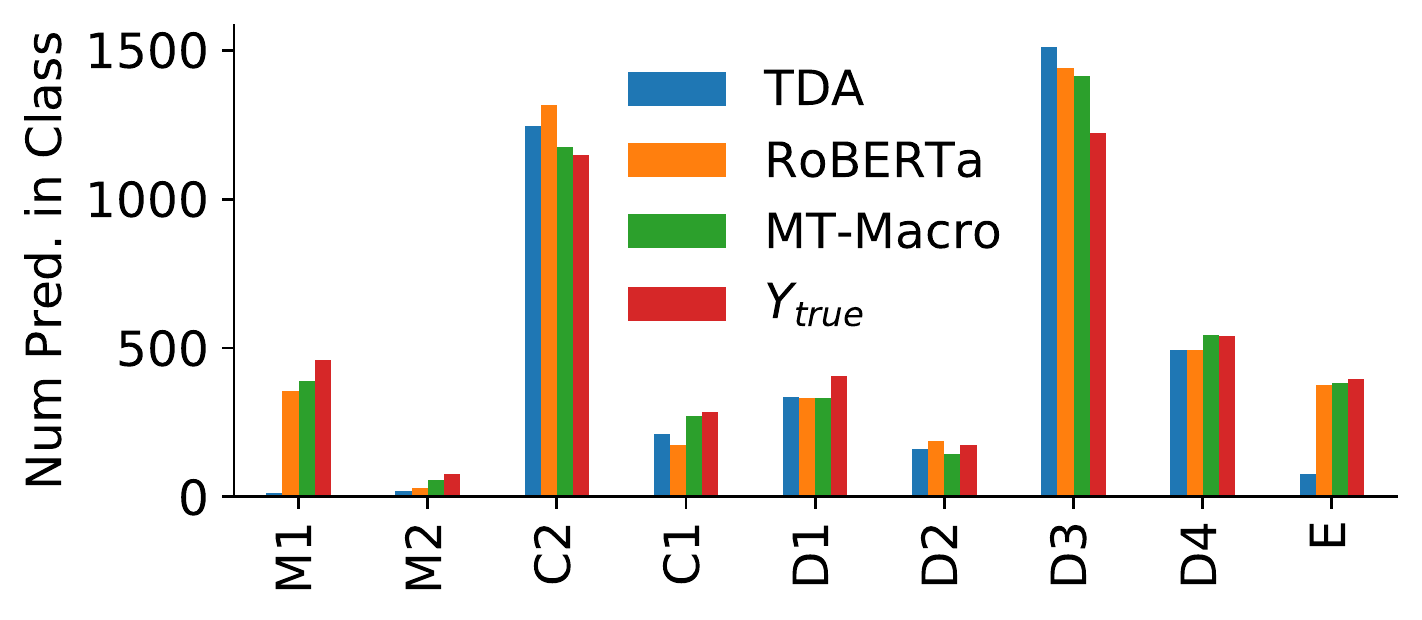}
    \caption{TDA over-predicts the better-represented classes (C2, D3) relative to $y_{true}$, and underpredicts the lesser represented classes (M1, M2, C1, D1). MT-Macro prediction rates are closer to $Y_{true}$. (If $C_{x}$ is the empirical distribution over class-predictions made by model $x$, then $D_{\text{KL}}(C_{\text{TDA}}||C_{Y_{true}})=.27$,  $D_{\text{KL}}(C_{\text{MT-Macro}}||C_{Y_{true}})=.01$).}
    \label{fig:num_predicted}
\end{figure}

\newcite{hyun2020class} show that, for class-imbalanced problems, regions of the data manifold that contain the underrepresented classes are poorly generalizable in data-augmented settings, resulting in general ambiguity in these regions. We show in Figure \ref{fig:num_predicted} that TDA over-predicts the overrepresented class.

\begin{table*}[t]
\centering 
\setlength\tabcolsep{1pt}
\renewcommand{\arraystretch}{0.8} 
\begin{tabular}{|l||lllll||lllllllllll||lllllllll||l|}
\hline
{} & \rot{Anecdote} & \rot{Assumption} & \rot{Common Ground} & \rot{Statistics} & \rot{Testimony} & \rot{Lede} & \rot{Main Event} & \rot{Consequence} & \rot{Previous Event} & \rot{Secondary Event} & \rot{Current Context} & \rot{Historical Event} & \rot{Evaluation} & \rot{Expectation} & \rot{Explanation} & \rot{Verbal Reaction} & \rot{Lede} & \rot{Main Event} & \rot{Consequence} & \rot{Previous Event} & \rot{Current Context} & \rot{Historical Event} & \rot{Evaluation} & \rot{Expectation} & \rot{Verbal React.} &  Supp. \\
\hline
Main Event       &             {} &               {} &                  {} &               {} &              {} &    \cca{3} &          \cca{4} &                {} &              \ccb{4} &                    {} &                    {} &                     {} &               {} &                {} &                {} &                    {} &    \cca{4} &               {} &                {} &                   {} &                    {} &                \ccb{4} &               {} &                {} &                    {} &     1782 \\
Consequence      &             {} &               {} &             \cca{5} &               {} &              {} &    \cca{4} &               {} &           \cca{4} &                   {} &               \cca{5} &                    {} &                     {} &               {} &                {} &                {} &                    {} &    \cca{3} &          \cca{5} &           \cca{5} &                   {} &                    {} &                     {} &               {} &                {} &                    {} &      387 \\
Previous Event   &        \cca{5} &               {} &             \cca{3} &          \cca{5} &              {} &         {} &               {} &                {} &              \cca{3} &                    {} &               \cca{7} &                     {} &               {} &           \ccb{4} &                {} &               \ccb{6} &         {} &               {} &                {} &              \cca{4} &               \cca{7} &                     {} &               {} &                {} &               \ccb{4} &     4486 \\
Current Context  &        \cca{4} &               {} &                  {} &          \cca{4} &              {} &         {} &          \cca{4} &                {} &              \cca{4} &                    {} &               \cca{6} &                     {} &               {} &           \ccb{5} &                {} &               \ccb{5} &         {} &          \cca{3} &                {} &              \cca{5} &                    {} &                     {} &          \ccb{4} &           \ccb{4} &                    {} &     1094 \\
Historical Event &             {} &               {} &                  {} &               {} &              {} &         {} &               {} &           \ccb{4} &              \cca{7} &                    {} &                    {} &                \cca{6} &               {} &                {} &                {} &                    {} &         {} &               {} &                {} &              \cca{5} &                    {} &                \cca{8} &               {} &                {} &                    {} &     1499 \\
Anecdotal Event  &             {} &               {} &                  {} &               {} &         \cca{3} &         {} &               {} &                {} &                   {} &                    {} &               \ccb{4} &                     {} &               {} &                {} &                {} &               \cca{6} &         {} &               {} &                {} &                   {} &               \ccb{4} &                     {} &               {} &                {} &               \cca{5} &      609 \\
Evaluation       &        \ccb{4} &               {} &             \ccb{5} &               {} &         \cca{5} &    \ccb{4} &          \ccb{4} &                {} &                   {} &               \ccb{5} &                    {} &                     {} &               {} &                {} &                {} &               \cca{5} &    \ccb{4} &          \ccb{6} &           \ccb{4} &              \ccb{4} &                    {} &                     {} &          \cca{3} &                {} &               \cca{6} &     4697 \\
Expectation      &        \ccb{4} &               {} &                  {} &               {} &         \cca{3} &         {} &               {} &           \cca{3} &              \ccb{5} &                    {} &               \ccb{4} &                \ccb{6} &               {} &           \cca{7} &                {} &               \cca{5} &         {} &               {} &                {} &              \ccb{5} &                    {} &                     {} &               {} &           \cca{7} &               \cca{4} &     1981 \\
\hline
\hline
{} & \multicolumn{5}{c||}{Argument.} & \multicolumn{11}{c||}{\spangh Dataset} & \multicolumn{9}{c||}{\fin Dataset} & {} \\
\hline
\end{tabular}
\caption{Spearman correlations between tags predicted with \nd head and Argumentation, \spangh and \fin heads. Note that the two Van Dijk datasets have high correlations between most tags that they have in common.}
\label{tbl:corr_arg_vd}
\end{table*}

\begin{table*}[t]
\centering
\setlength\tabcolsep{3.5pt}
\renewcommand{\arraystretch}{0.8} 
\begin{tabular}{|l||llllllllllll||lllll||l|}
\hline
{} & \rot{Elaboration} & \rot{Joint} & \rot{Topic Change} & \rot{Attribution} & \rot{Contrast} & \rot{Explanation} & \rot{Background} & \rot{Evaluation} & \rot{Summary} & \rot{Cause} & \rot{Topic Comment} & \rot{Temporal} & \rot{Temporal} & \rot{Asynchronous} & \rot{Precedence} & \rot{Synchrony} & \rot{Succession} &  Support \\
\hline
Main Event       &                {} &          {} &                 {} &                {} &             {} &                {} &          \ccb{4} &               {} &            {} &          {} &                  {} &             {} &             {} &                 {} &               {} &              {} &               {} &     1782 \\
Consequence      &                {} &          {} &            \cca{4} &                {} &             {} &           \ccb{4} &               {} &               {} &            {} &          {} &                  {} &             {} &        \cca{2} &            \cca{2} &               {} &         \cca{2} &          \cca{3} &      387 \\
Previous Event   &                {} &          {} &                 {} &           \ccb{4} &             {} &                {} &          \cca{4} &          \ccb{6} &            {} &          {} &             \ccb{5} &             {} &             {} &                 {} &               {} &              {} &               {} &     4486 \\
Current Context  &                {} &          {} &                 {} &                {} &             {} &                {} &               {} &               {} &            {} &          {} &                  {} &             {} &             {} &                 {} &               {} &              {} &               {} &     1094 \\
Historical Event &                {} &          {} &                 {} &                {} &             {} &                {} &          \cca{3} &               {} &            {} &          {} &                  {} &        \cca{3} &        \cca{2} &            \cca{3} &          \cca{3} &              {} &          \cca{2} &     1499 \\
Anecdotal Event  &                {} &          {} &                 {} &           \cca{3} &        \cca{3} &           \cca{4} &               {} &          \cca{5} &            {} &     \cca{3} &             \cca{4} &             {} &             {} &                 {} &               {} &              {} &               {} &      609 \\
Evaluation       &                {} &          {} &                 {} &           \cca{4} &        \cca{5} &           \cca{4} &               {} &          \cca{6} &            {} &          {} &                  {} &             {} &        \ccb{3} &            \ccb{4} &          \ccb{4} &         \ccb{3} &          \ccb{4} &     4697 \\
Expectation      &                {} &          {} &                 {} &           \cca{3} &             {} &                {} &               {} &          \cca{4} &            {} &          {} &                  {} &        \ccb{4} &             {} &                 {} &               {} &              {} &               {} &     1981 \\
\hline
{} & \multicolumn{12}{c||}{RST Dataset} & \multicolumn{5}{c||}{PDTB-$t$ Dataset} & {}\\
\hline
\end{tabular}
\caption{Spearman correlation between tags predicted with \nd head and RST head and PDTB-$t$ head, on the Evaluation split of \nd. Note that PDTB-$t$ relations, which tend to be temporally-based, have a positive correlation with \textit{Consequence} and \textit{Historical Event} tags, which are both defined in temporal relation to the \textit{Main Event} tag.}
\label{tbl:corr_pdtb_rst}
\end{table*}

Multitask learning can help learn part of the data manifold where underrepresented class exists by learning signal from a class which is correlated. Tables \ref{tbl:corr_arg_vd} and \ref{tbl:corr_pdtb_rst} show the correlation between class labels predicted by our multitask model on the same dataset using different heads. For example, for a set of sentences $X$, there is a $.4$ correlation between those tagged \textit{Background} by the RST head, and those tagged \textit{Previous Event} by the \nd head.

\begin{figure}[t]
    \centering
    \includegraphics[width=\linewidth]{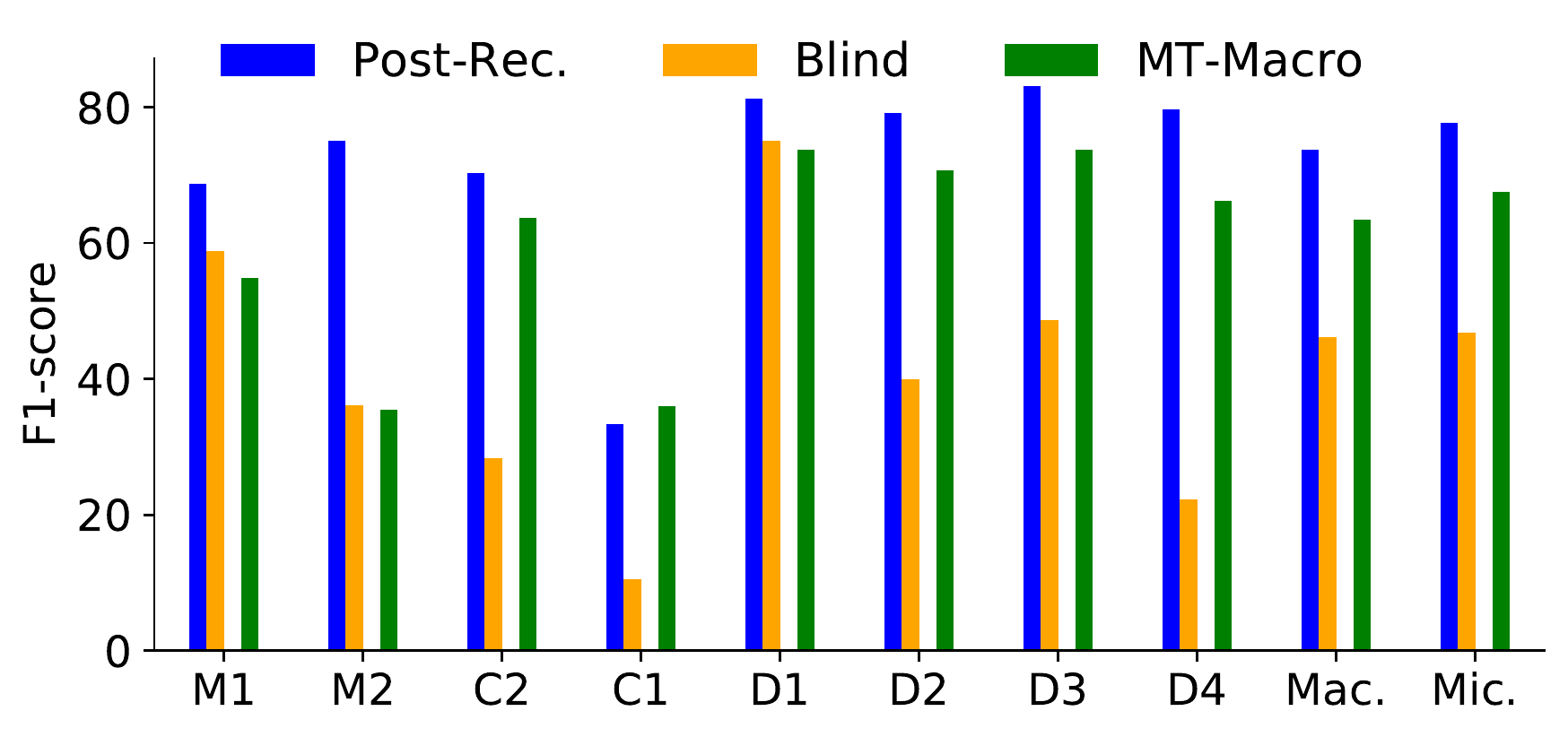}
    \caption{Our agreement with \cite{choubey-etal-2020-discourse} on a small $30$ article sample from \nd. \textbf{Blind} shows our agreement after reading their annotation guidelines and practicing, but not observing their results. \textbf{Post-Rec.} (Post-Reconciliation) shows our agreement after observing their annotations on the documents we annotated. \textbf{MT-Macro} is one of our top models.}
    \label{fig:human_perf}
\end{figure}

Table \ref{tbl:corr_arg_vd} provides a sanity check: the Van Dijk datasets largely agree on the tags that share similar definitions. For example, there is a strong correlation between sentences tagged \textit{Main Event} by the \nd head and those tagged \textit{Main Event} by the \spangh head. 

However, what is most interesting are the strong correlations existing between underrepresented classes in the \nd dataset. Classes \textit{Consequence} and \textit{Anecdotal Event} are two of the lowest-support classes, yet they each have strong correlations with tags in every other dataset. For example, a \textit{Consequence} sentence, which is defined in part due to its temporal relation to a \textit{Main Event} sentence, is correlated with temporal tags in the PDTB-$t$ dataset. Likewise, \textit{Anecdotal Event} is correlated with \textit{Testimony} in the Argumentation dataset.


TDA serves as an ablation study.
A counterargument to our claims on our Multitask setup is that, by incorporating additional tasks and additional datasets, we simply expose the model to more data. As TDA shows, however, this is not the case, as performance drops, in TDA's case significantly, when we introduce more data.\footnote{One approach we can consider for TDA is to generate more augmentations for underrepresented classes. However, since we are modeling sequential data, this is generally not possible to do for all underrepresented tags.}

We provide additional information on the difficulty of this task in Figure \ref{fig:human_perf} by asking additional expert annotators to label our data. Our annotators sampled 30 documents from \nd, read \newcite{choubey-etal-2020-discourse}'s annotation guidelines and practiced on a few trial examples. Then they annotated all 30 documents. Annotations made in this pass, the \textbf{Blind} pass, had significantly lower accuracy across categories than our best model. Then, however, our annotators observed \newcite{choubey-etal-2020-discourse}'s original tags on the 30 blind-tagged articles, discussed, and changed where necessary. Surprisingly, even in this pass, the \textbf{Post-Reconciliation} pass, our annotators rarely had more than 80\% F1-score agreement with \newcite{choubey-etal-2020-discourse}'s published tags. 

Thus, Van Dijk labeling task might face an inherent level of legitimate disagreement, which \textbf{MT-Macro} seems to be approaching. However, there are two classes, M1 and M2, where \textbf{MT-Macro} underperformed even the \textbf{Blind} annotation. For these classes, at least, we expect that there is further room for modeling improvement through: (1) annotating more data, (2) incorporating more auxiliary tasks in the multitask setup (3) learning from unlabeled data, using an algorithm like MixMatch \cite{mixmatch} or unsupervised data augmentation \cite{xie2019unsupervised} along with our supervised tasks.



\section{Related Work}

Most state-of-the-art research in discourse analysis has focused on classifying the discourse relations between pairs of clauses, as is practice in the Penn Discourse Treebank (PDTB) \cite{prasad2008penn} and Rhetorical Structure Theory (RST) dataset \cite{carlson2003building}. Corpora and methods have been developed to predict explicit discourse connectives \cite{miltsakaki-etal-2004-annotating,lin2009recognizing,das2018constructing,malmi2017automatic,wang-etal-2018-toward} as well as implicit discourse relations  \cite{rutherford2016robust,liu2016implicit,lan2017multi,lei2017swim}.  \citet{choubey-etal-2020-discourse} built a news article corpus where each sentence was tagged with a discourse label defined in Van Dijk schema \cite{van2013news}. 

Since discourse analysis has limited resources, some work has explored multitask framework to learn from more than one discourse corpus. \citet{liu2016implicit} propose a CNN based multitask model and \citet{lan2017multi} propose an attention-based multitask model to learn implicit relations in PDTB and RST. The main difference in our work is the coverage and flexibility of our framework. This work is able to learn both explicit and implicit discourse relations, both multilabel and multiclass tasks, and both labeled data and non-labeled data in one framework, which makes it possible to fully utilize classic corpora like PDTB and RST as well as recent corpora developed in Van Dijk schema.

\citet{ruder2017overview} gives a good overview of multitask learning in NLP more broadly. A major early work by \citet{collobert2008unified} uses a single CNN architecture to jointly learn $6$ different NLP tasks, ranging from supervised low-level syntactic tasks (e.g. \textit{Part-of-Speech Tagging}) to higher-level semantic tasks (e.g. \textit{Semantic Role Labeling}) as well as unsupervised tasks (e.g. \textit{Language Modeling}). They find that a multitask improves performance in semantic role labeling. Our work differs in several key aspects: (1) we are primarily concerned with sentence-level tasks, whereas \citeauthor{collobert2008unified} perform word-level tasks; (2) we consider a softer approach to task inclusion and use different weighting schemes for the tasks in our formulation; (3) we perform a deeper analysis of why multitask helps, including examining inter-task prediction-correlations and class-imbalance.

Another broad domain of multitask learning in NLP lies within machine translation, the canonical example being \citet{aharoni2019massively}. In this work, authors jointly train a translation model between hundreds of language-pairs and find that low-resource tasks benefit especially. A different direction for multilingual modeling has been to use resources from one language to perform tasks in another. For many NLP tasks, for example, Information Extraction \cite{wiedemann2018multilingual,neveol2017clef,poibeau2012multi}, Event Detection \cite{liu2018event,agerri2016multilingual,lejeune2015multilingual}, Part-of-Speech tagging \cite{plank2016multilingual,naseem2009multilingual}, and even Discourse Analysis \cite{liu2020multilingual}, the largest tagged datasets available are primarily in English, and researchers have trained classifiers in a multilingual setting that either translate resources into the target language, translate the target language into the source language, or learn a joint multilingual space. We are primarily concerned with labeling discourse-level annotations on English sentences, however, we may benefit from multilingual discourse datasets. 

\section{Conclusion}

We have shown a state-of-the-art improvement of $7\%$ Micro F1-score above baseline, from 62.8\% F1-score to 67.7\% F1-score, for discourse tagging on the \textit{NewsDiscourse} dataset, the largest dataset currently available focusing on sentence-level Van Dijk discourse tagging. This dataset has a number of challenges: distinctions between Van Dijk discourse tags are based on a number of complex attributes, which our baseline models showed high confusion (Appendix Figure~\ref{sfig:baseline_confusion_mat}), for example, temporal relations between different sentences or the presence or absence of an event). Additionally, this dataset is class-imbalanced, with the overrepresented classes being, on average, $3$ times more likely than the underrepresented classes. 

We showed that a multitask approach could be especially helpful in this circumstance, improving performance for underrepresented tags more than overrepresented tags. A possible reason, we show, is the high correlations we observe between tag predictions between tasks, indicating that auxiliary tasks are giving signal to underrepresented tags in our primary task. This includes high correlations observed in a novel dataset that we introduce based on the same schema with some minor alterations. This raises the additional benefit that our multitask approach can reconcile different datasets with slightly different schema, allowing NLP researchers not to ``waste'' valuable tagging work.

Finally, we perform a comparative analysis of other strategies proposed in the literature for dealing with small datasets or class-imbalanced problems: specifically, changing loss functions, hierarchical classification and CRF-sequential modeling. We show in exhaustive experiments that these approaches do not help us improve above baseline. These negative experiments include important analysis for future researchers, and provide a powerful justification for the necessity of our multitask approach.

\bibliographystyle{acl_natbib}
\bibliography{custom}

\appendix
\section{Appendices Overview} 

\begin{table*}
\centering
\begin{tabular}{l||lllllll||r|r|}
{} & \rot{\argumentation} &  \rot{\spangh} & \rot{\fin} &       \rot{RST} &      \rot{PDTB-$t$} & \rot{KBP} &    \textbf{Tag F1-Score} & {(MT-Micro)}\\
\hline
Main Event       &      \ccd{28} &        {} &        {} &  \ccd{19} &        {} &  {} &  58.37 &    (54.91) \\
Consequence      &            {} &        {} &  \ccd{18} &  \ccd{27} &        {} &  {} &  40.00 &    (35.48) \\
Previous Event   &      \ccd{30} &        {} &        {} &  \ccd{10} &  \ccd{10} &  {} &  67.06 &    (63.76) \\
Current Context  &      \ccd{27} &   \ccd{9} &   \ccd{9} &        {} &        {} &  {} &  38.75 &    (35.94) \\
Historical Event &            {} &        {} &  \ccd{18} &  \ccd{27} &        {} &  {} &  77.02 &    (73.71) \\
Anecdotal Event  &       \ccd{9} &   \ccd{9} &   \ccd{9} &   \ccd{9} &   \ccd{9} &  {} &  75.84 &    (70.73) \\
Evaluation       &            {} &  \ccd{18} &   \ccd{9} &        {} &  \ccd{18} &  {} &  74.78 &    (73.71) \\
Expectation      &      \ccd{10} &        {} &        {} &  \ccd{10} &  \ccd{30} &  {} &  68.94 &    (66.26) \\
\hline
\end{tabular}
\caption{Maximum multitask weighting, $\alpha$, by tag, for secondary datasets. \textbf{Tag F1-score} shows the maximum F1-score for the tag, and the left columns show the $\alpha$ that achieves this weighting. Right-most column is shown simply for comparison. Note that PDTB-$t$ contributes most to \textit{Expectation}, while Argumentation contributions most to \textit{Main Event, Previous Event} and \textit{Current Context}.}
\label{tbl:ind_tags_alphs}
\end{table*}

The appendices convey two broad areas of analysis: (1) Additional explanatory information for our multitask setup and (2) Negative Experiments and Results.

Appendix \ref{app:datasets} contains more information on the datasets used. Appendix \ref{app:confusion_matrices} and \ref{app:multitask_dataset_contributions} contain explanatory analysis. Appendix \ref{app:confusion_matrices} shows that our multitask setup is reducing confusion between several important pairs of tags, and Appendix \ref{app:multitask_dataset_contributions} shows, for each tag, which $\alpha$-weighting across tasks yields the highest score.

Appendix \ref{app:neg_results}
provides more information about the negative results we obtained throughout our research and the explorations we performed. Appendix \ref{app:neg_results} specifically contains details about the additional experiments we ran and the results we obtained. We believe that it is important to publish about negative results, to help fight against publication bias \cite{easterbrook1991publication} and to help other researchers considering similar techniques. Where possible, we conducted explorations to understand why such results were negative, and what hyperparameters might be tuned to produce a positive results.

\section{Dataset Processing}
\label{app:datasets}

We summarize the tag-set in each of the datasets we used in Table \ref{tab:dataset_tagsets}, and in Table \ref{tab:RST_mapper}, we show the heuristic mapping scheme that we developed to reduce the dimensionality of the RST dataset.

\begin{table*}[t]
    \centering
    \begin{tabular}{|p{3cm}|p{12cm}|}
        \hline
        \textbf{Schema Name} & \textbf{Tagset} \\
        \hline
        \hline
         Van Dijk Schema &  $\{$ \textit{Lede},
        \textit{Main Event} (\textbf{M1}), \textit{Consequence} (\textbf{M2}), \textit{Circumstances} (\textbf{C1}), \textit{Previous Event} (\textbf{C2}), \textit{Historical Event} (\textbf{D1}), \textit{Expectation} \textbf{(D4)}, \textit{Evaluation} (\textbf{D3}), \textit{Verbal Reaction} $\}$ \\
        \hline
        \nd & Van Dijk $\oplus$ $\{$ \textit{Anecdotal Event} (\textbf{D2}) $\}$\\
        \hline
        \spangh & Van Dijk $\oplus$ $\{$ \textit{Explanation}, \textit{Secondary Event} $\}$ \\
        \hline
        \hline
        Argumentation & $\{$ \textit{Anecdote}, \textit{Assumption}, \textit{Common-Ground}, \textit{Statistics}, \textit{Testimony} $\}$ \\
        \hline
        \hline
        Penn Discourse Treebank & $\{$ \textit{Temporal}, \textit{Asynchronous}, \textit{Precedence}, \textit{Synchrony}, \textit{Succession} $\}$ \\
        \hline
        Rhetorical Structure Theory & $\{$ \textit{Elaboration}, \textit{Joint},  \textit{Topic Change},  \textit{Attribution}, \textit{Contrast}, \textit{Explanation}, \textit{Background},
        \textit{Evaluation}, \textit{Summary}, \textit{Cause}, \textit{Topic-Comment}, \textit{Temporal} $\}$ \\
        \hline
        KBP Event Nugget & $\{$ \textit{Actual Event}, \textit{Generic Event}, \textit{Event Mention}, \textit{Other} $\}$\\
        \hline
    \end{tabular}
    \caption{Overview of the tagsets for each of the datasets used.}
    \label{tab:dataset_tagsets}
\end{table*}

\begin{table*}[t]
    \centering
\begin{tabular}{|p{2cm}|p{13cm}|}
\hline
RST Tag-Class & RST Tags in Class \\
\hline
\hline
Attribution   & Attribution, Attribution-negative \\
\hline
Evaluation    & Evaluation, Interpretation, Conclusion, Comment \\
\hline
Background    & Background, Circumstance \\
\hline
Explanation   & Evidence, Reason, Explanation-argumentative \\
\hline
Cause         & Cause, Result, Consequence, Cause-result \\
\hline
Joint         & List, Disjunction \\
\hline
Comparison    & Comparison, Preference, Analogy, Proportion \\
\hline
Manner-Means  & Manner, Mean, Means \\
\hline
Condition     & Condition, Hypothetical, Contingency, Otherwise \\
\hline
Topic-Comment & Topic-comment, Problem-solution, Comment-topic, Rhetorical-question, Question-answer \\
\hline
Contrast      & Contrast, Concession, Antithesis \\
\hline
Summary       & Summary, Restatement, Statement-response \\
\hline
Elaboration   & Elaboration-additional, Elaboration-general-specific, Elaboration-set-member, Example, 
Definition, Elaboration-object-attribute, Elaboration-part-whole, Elaboration-process-step \\
\hline
Temporal      & Temporal-before, Temporal-after, Temporal-same-time, Sequence, Inverted-sequence \\
\hline
Enablement    & Purpose, Enablement \\
\hline
Topic Change  & Topic-shift, Topic-drift \\
\hline
\end{tabular}
    \caption{The mapping we developed to reduce dimensionality of the RST Treebank. The left column shows the tag-class which we ended up using for classification and the right column shows the RST tags that we mapped to that category. Tag-mapping was done heuristically.}
    \label{tab:RST_mapper}
\end{table*}

\section{Confusion Matrices}
\label{app:confusion_matrices}

\begin{figure}[t]
\begin{subfigure}{.5\textwidth}
  \centering
  \includegraphics[width=\linewidth]{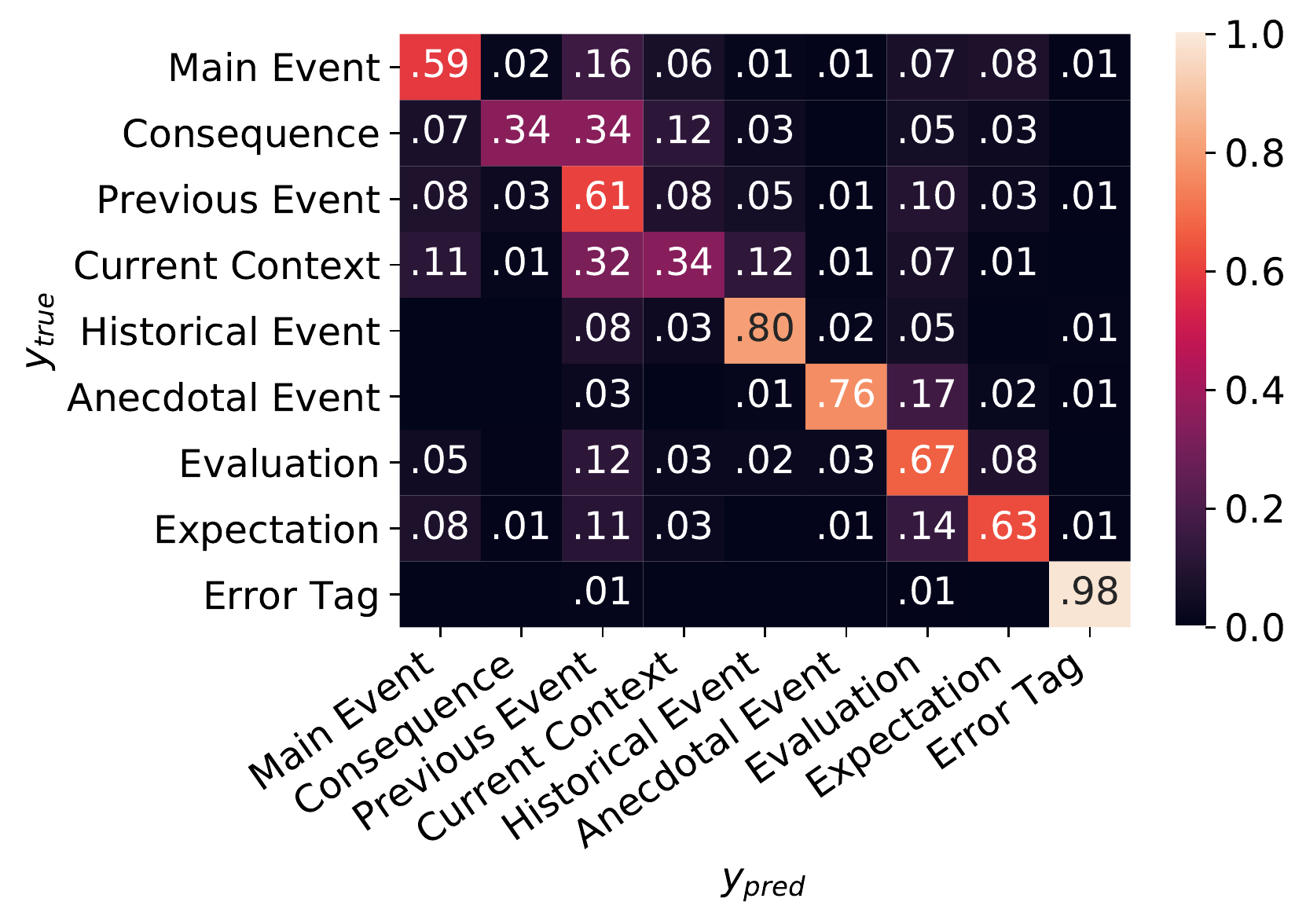}
  \caption{\textbf{Baseline } confusion matrix (for RoBERTa +EmbAug.) Major classes of confusion are: (a)\textbf{Temporal}, ex. between \textit{Consequence} and \textit{Previous Event} (b) \textbf{Semantic}, ex. between \textit{Current Context} and \textit{Previous Event}.}
  \label{sfig:baseline_confusion_mat}
\end{subfigure}
\begin{subfigure}{.5\textwidth}
  \centering
  \includegraphics[width=\linewidth]{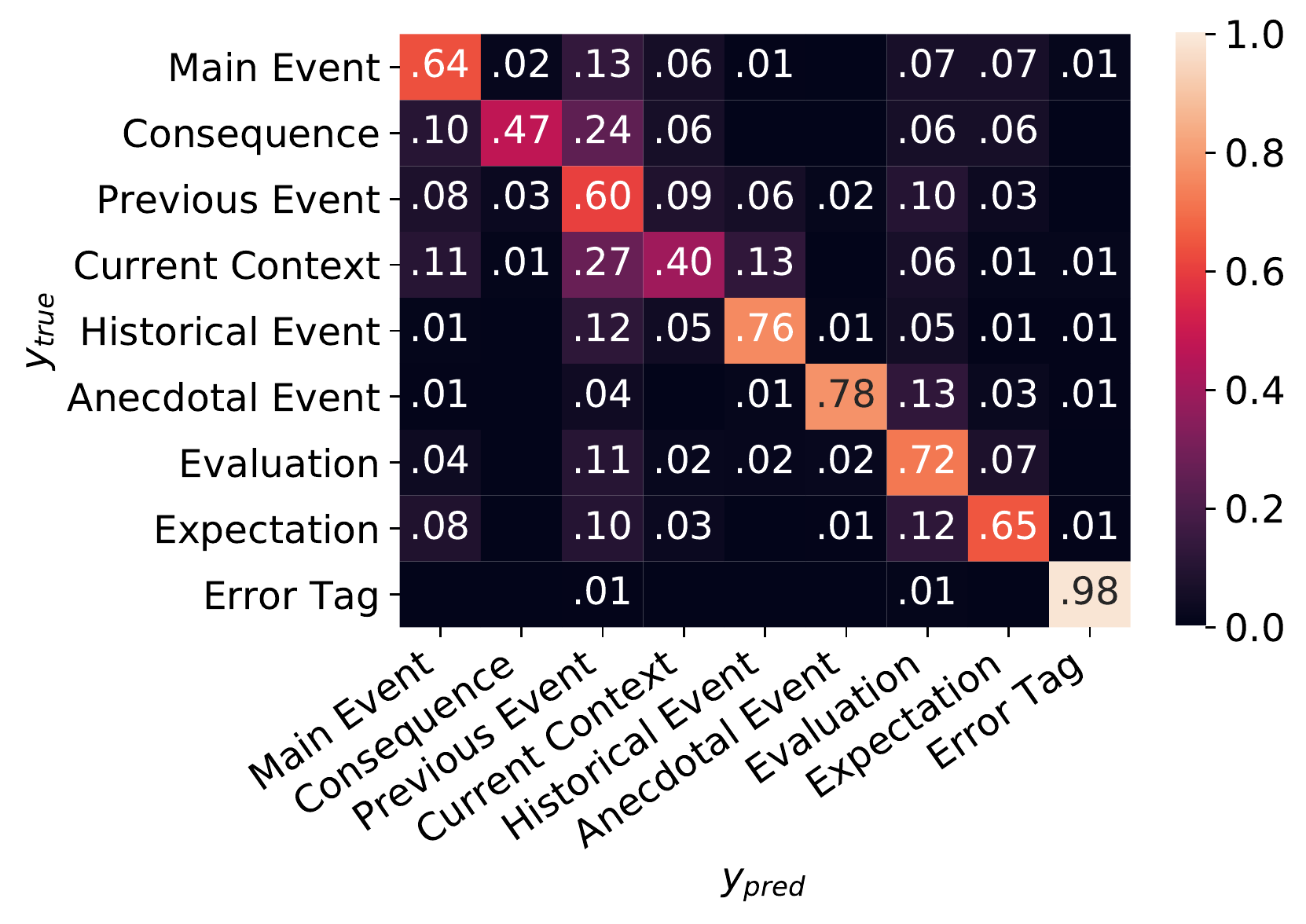}
  \caption{\textbf{MT-Macro} confusion matrix. We see a significant reduction in uncertainty for both semantic and temporal confusions.}
  \label{sfig:multitask_confusion_mat}
\end{subfigure}
\caption{Confusion Matrices for Baseline RoBERTa compared with MT-Macro.}
\label{fig:fig}
\end{figure}

Based on the confusion matrix shown in Figure \ref{sfig:baseline_confusion_mat}, we identify two important classes of error: \textbf{Semantic} error and \textbf{Temporal} error. These two types of error can be illustrated by the two classes with the highest confusion, \textit{Consequence} and \textit{Current Context} (these classes of error are also evident in other confusions). 

\textbf{Semantically}, a lot of tags differ based on whether a specific discourse span contains an event. For instance, both \textit{Current Context} and \textit{Previous Event} describe the lead-up to a \textit{Main Event}, however \textit{Previous Event} contains the literal description of an event, while \textit{Current Context} does not. (A similar confusion can be seen between \textit{Anecdotal Event} and \textit{Evaluation}.) We hypothesized, thus, that adding an Event-Nugget dataset, or a dataset specifically focused on identifying events, would help us with these confusions. However, that was not observed, as KBP decreased the performance of our multitask approach.

\textbf{Temporally}, many tags are defined based on the temporal relation of events in discourse spans relative to the \textit{Main Event} of an article. For example, \textit{Previous Events}, \textit{Historical Events} and \textit{Current Contexts} happen before the \textit{Main Event}, while \textit{Consequences} and \textit{Expectations} happen after. The major confusion occurring between \textit{Previous Event} and \textit{Consequence} is an example of a temporal confusion: sentences describing an event happening \underline{after} the \textit{Main Event} are misinterpreted to be \underline{before}. (The confusion between \textit{Expectation} and \textit{Previous Event} is another example of such a confusion.) To address this confusion, we considered adding a temporal-relation-based dataset, like MATRES \cite{ning2018multi}, but instead filtered down the PDTB to include temporal relations. As can be shown in Table \ref{tbl:corr_arg_vd}, PDTB-$t$ is positively correlated with \textit{Consequence}, and as shown in Table \ref{tbl:ind_tags_alphs}, PDTB-$t$ contributes to temporal tags like \textit{Previous Event} and \textit{Expectation}.

As shown in Figure \ref{sfig:multitask_confusion_mat}, the addition of the multitask datasets decreased confusion in these two main classes, reducing \textbf{Temporal} confusion between \textit{Consequence} and \textit{Previous Event}, and \textbf{Semantic} confusion, between \textit{Current Context} and \textit{Previous Event}, among other pairs of tags.

\section{Interrogating Multitask Dataset Contributions}
\label{app:multitask_dataset_contributions}

In the main body of the paper, we interpreted the effects of the multitask setup by examining the overall increase in performance (Figure \ref{fig:multitask_coefs}), the regressive effects of each dataset (Table \ref{tbl:lin_reg}) and the correlations between tag-predictions (Tables \ref{tbl:corr_arg_vd}, \ref{tbl:corr_pdtb_rst}). Another way to examine the contributions of each task is to analyze which combination of datasets results in the highest F1-score for each tag.

In Table \ref{tbl:ind_tags_alphs}, we show the $\alpha$-weighting that results in the optimal F1-score for each tag. This gives us not only a sense of which datasets are important for that tag, but how much of an improvement we can seek over the baseline MT-Micro. For instance, a strong $.3$ weight for PDTB-$t$ increases the performance for the \textit{Expectation} tag and a strong $.27$ weight for RST increases the performance of the \textit{Historical Event} tag. This is possibly because both the \textit{Expectation} tag and the \textit{Historical Event} tag describes events either far in the future or far in the past relative to the \textit{Main Event}, and both PDTB-$t$ and RST contain information about temporal relations. Interestingly, and perhaps conversely, a strong $\alpha$-weighting for the \argumentation dataset ($>.25$) increases performance for \textit{Main Event}, \textit{Previous Event}, and \textit{Current Context}. This set of tags might seem counterintuitive, since they are all dealing with factual statements and events, and by definition contain less commentary and opinion than tags like \textit{Expectation} and \textit{Evaluation}. However, if we cross-reference this table, Table \ref{tbl:ind_tags_alphs}, with Table \ref{tbl:corr_arg_vd}, we can see strong positive correlations between these tags and \argumentation tags like \textit{Common Ground}, \textit{Statistics} and \textit{Anecdote}. (It's surprising that \argumentation's \textit{Anecdote} tag does not correlate with \nd's \textit{Anecdotal Event} tag, but perhaps the definitions are different enough that, despite the semantic similarity between the labels, they are in fact capturing different phenomena.)

\section{Additional Negative Results}
\label{app:neg_results}
In this section, we describe additional experiments. 
They do not improve the accuracy of our task, but we have not done the necessary analysis to determine \textit{why}, and what their shortcomings tell us about the nature of our problem. We hope, though, that by sharing our exploration in this Appendix, we might inspire researchers working with similar tasks to consider these methods, or advancements of them. Table \ref{tab:negative_results_uninformative} shows the results of the experiments described in this section.
 
\subsection{Sentence Embedding Variations}
\label{app:sent_emb}

There are, as of this writing, three different transformer-based sentence-embedding techniques in the literature: Sentence-BERT and Sentence Weighted-BERT \cite{reimers2019sentence}, and SBERT-WK \cite{wang2020sbert}. Sentence-BERT trains a Siamese network to directly update the <s> token. Sentence Weighted BERT learns a weight function for the word embeddings in a sentence. SBERT-WK proposes heuristics for combining the word embeddings to generate a sentence-embedding.

None of the sentence-embedding variations yielded any improvement above the RoBERTa <s> token. It's possible that these models, which were designed and trained for NLI tasks, do not generalize well to discourse tasks. Additionally, we test the following baselines: the CLS token from BERT-base embeddings, and generating sentence-embeddings using self-attention on Elmo word-embeddings, as described in \cite{choubey-etal-2020-discourse}. These baselines show no improvement above RoBERTa. We see a strong need for a general pretrained sentence embedding model that can transfer well across tasks. We envision a sort of masked-sentence model, instead of a masked-word model, leaving this open to future research.

\subsection{Supervised Head Variations}
\label{app:supervised_head_variations}
\subsubsection{Classification Task Variations}
\label{app:classification_variations}

For variations on the classification task, we consider using a Conditional Random Field layer instead of a simple FFNN layer, which has been shown to improve results \cite{li2019discourse}. However, we do not see an improvement in this case, possibly because the Bi-LSTM layer prior to classification was already inducing sequential information to be shared.

We also consider taking a hierarchical classification approach. Inspired by \cite{silva2017improving}, we construct $C$ clusters of semantically-related labels such that each class falls into one cluster\footnote{Semantic-relatedness is given a-prior by the tag definitions, for more information, see \cite{yarlott2018identifying, choubey-etal-2020-discourse}}. We construct variables from each $y_i$: $\hat{y_i}^{(c)}$,  $\hat{y_i}^{(c_0)}...\hat{y_i}^{(c_k)}$:
\begin{gather*}
    \hat{y_i}^{(c)} = \{\mathbbm{1}(y_i \in  \text{cluster } j)\}_{j=1}^C\\
    \hat{y_i}^{(c_0)} = \{\mathbbm{1}(y_i = l)\}_{l=1}^{N_{c_0}}\\
    ...\\
    \hat{y_i}^{(c_k)} = \{\mathbbm{1}(y_i = l)\}_{l=N_{c_0}+...+N_{c_{k-1}}}^{N_{c_0} + ... + N_{c_k}}
\end{gather*}
where $C$ is the number of clusters of semantically-related classes and $L$ is the original number of labels. We try modeling these variables two ways. (1) As a 2-level hierarchy, where the top-level is one task and each sublayer is a separate task or (2) as a multilabel classification task of $\hat{y_i}$, where $\hat{y_i} = \hat{y_i}^{(c)} \oplus \hat{y_i}^{(c_0)} \oplus ... \oplus \hat{y_i}^{(c_k)}$. 

Our hierarchical classification shows no improvement above vanilla multiclass classification. It's possible that the transformer architecture is already learning the label hierarchy implicitly, and the information we try to pass in by structuring the output space does not improve the prediction.

\subsubsection{Loss Variations}
\label{app:Loss Variations}

\begin{table}[h]
    \centering
    \begin{tabular}{|l|rr|}
    \hline
    Method &  Mac. &  Mic. \\
    \hline
    GDL   &      55.45 &         64.41 \\
    $\text{GDL}^{(2)}$ &      49.90 &         62.82 \\
    GADL &      29.39 &         41.97 \\
    \hline 
    \hline 
    (MT-Micro) & \textit{61.89} & \textit{67.70} \\
    \hline
    \end{tabular}
    \caption{Macro-F1 (Mac.) and Micro F1 (Mic.) scores for variations of Multiclass Dice Loss. DL: Vanilla Dice Loss, $DL^{(2)}$: the Square Form of Dice Loss, $ADL$: self-adjusting dice loss \cite{li2019dice}. Multiclass generalized as in \cite{sudre2017generalised}.}
    \label{tbl:dice_loss}
\end{table}

\begin{table*}[t]
    \centering
    \begin{tabular}{|l||r|r|r|r|r|r|r|r|r||r|r|}
    \hline
    {} &           M1 &     M2 &    C2 &     C1 &     D1 &     D2 &     D3 &     D4 &      E &  M. &  W. \\
    \hline
    SBERT &        52.0 &  11.2 &  61.7 &  31.1 &  67.9 &  43.1 &  69.9 &  64.9 &  96.6 &  55.39 &     63.38 \\
    +Frozen &      54.8 &  19.3 &  62.6 &  29.9 &  70.2 &  53.5 &  70.0 &  61.8 &  96.2 &  57.59 &     64.14 \\
    +EmbAug &      54.6 &  25.0 &  62.8 &  33.0 &  69.8 &  45.7 &  71.9 &  65.2 &  95.7 &  58.20 &     64.95 \\
    \hline
    SWBERT &       51.3 &  14.5 &  61.3 &  30.2 &  70.1 &  55.1 &  71.2 &  64.3 &  97.0 &  57.23 &     64.14 \\
    +Frozen &      52.4 &  20.6 &  62.6 &  31.5 &  68.7 &  61.1 &  73.9 &  66.0 &  95.9 &  59.17 &     65.62 \\
    +EmbAug &      52.2 &  12.0 &  64.6 &  31.7 &  72.2 &  50.0 &  73.0 &  66.8 &  96.7 &  57.68 &     65.79 \\
    \hline
    Hier. &        47.5 &  0.0 &   59.4 &  24.3 &  68.3 &  66.0 &  71.6 &  63.8 &  91.3 &  54.68 &     62.51 \\
    Dice &         55.4 &  18.5 &  63.7 &  29.5 &  70.8 &  25.2 &  72.9 &  64.2 &  95.6 &  55.09 &     64.41 \\
    CRF &          54.6 &  16.4 &  62.8 &  30.0 &  70.1 &  65.5 &  72.3 &  64.2 &  96.2 &  59.13 &     65.43 \\
    \hline
    \hline
    \textit{MT-Mic} &  \textit{55.35} &  \textit{25.0} & \textit{67.06} &  \textit{32.78} &  \textit{72.5} &  \textit{68.88} &  \textit{73.63} &  \textit{65.8} & \textit{96.0} & \textit{61.89} & \textit{67.70} \\
    \hline
    \end{tabular}
    \caption{\textbf{Negative Results:} We show the results of experiments and manipulations that did not increase the accuracy of our model. For all variations that we report, we report the maximum score observed under an array of hyperparameter settings. Except for SBERT and SWBERT, which are shown, all of these tasks include +Freezing and +EmbAug, as shown in Table \ref{tbl:embedding_augmentations}.}
    \label{tab:negative_results_uninformative}
\end{table*}

For variations on the loss, we consider losses other than a vanilla Cross-Entropy loss for the multiclass tasks and Binary Cross-Entropy loss for the multilabel tasks. Specifically, we experiment with variations of Dice Loss for the multiclass tasks, which has been proposed for class-imbalanced classification problems in computer vision \cite{milletari2016v} and NLP \cite{li2019dice}. Dice Loss seeks to directly optimize F1-score. A differentiable F1-based loss function can be derived by noting that: for a model making predictions on a single datapoint, $(x_i, y_i)$, the precision of the prediction is Prec$(x_i) = p(y_i = 1 | x_i) = p_i$, the probability given by the model, and the recall of the model is the ground truth of that datapoint, Recall$(x_i) = y_i$.

\begin{gather*}
    \text{F1}(x_i) = \frac{2\text{Prec}(x_i) \times \text{Recall}(x_i)}{\text{Prec}(x_i) + \text{Recall}(x_i)}\\
    \text{Dice Score}(x_i) = \frac{2p_{i,1} y_{i,1}}{p_{i,1} + y_{i,1}}
\end{gather*}

Across an entire dataset, Binary Dice Loss can be expressed as:

\begin{equation}
DL(X) = 1 - \frac{2 \sum_i p_{i,1} y_{i,1} + N \gamma}{\sum_i p_{i,1} + \sum_i y_{i,1} + N \gamma}
\end{equation}

where $\gamma$ is a hyperparameter set (typically $\gamma = 1$) to ensure that negative examples ($y_i = 0$) also contribute to the loss. Binary Dice loss can also be expressed in the square form \cite{milletari2016v}:

\begin{equation}
    DL^{(2)}(X) = 1 - \frac{2 \sum_i p_{i,1} y_{i,1}  + N \gamma}{\sum_i p_{i,1}^2 + \sum_i y_{i,1}^2 + N \gamma}
\end{equation}

Additionally, \cite{li2019dice} proposed a self-adjusting Binary Dice Loss (ADL) by multiplying $p_i$ by $(1-p_i)$ to downweight ``easy'' examples, or examples where $p_i$ is close to $0$ or 1:

\begin{equation}
    ADL(X) = 1 - \frac{2(1 - p_{i,1})p_{i,1} y_{i,1} + \gamma}{(1 - p_{i,1})p_{i,1} + y_{i,1} + \gamma}
\end{equation}

A multiclass Dice Loss for $k$ classes can be derived either through macro-averaging, micro-averaging, or a squared sum introduced by \cite{sudre2017generalised}:

\begin{gather}
    \text{GDL}(X) = \sum_{j=1}^{k} \frac{1}{N_j^2} * DL(p_j, y_j)
\end{gather}

As shown in Table \ref{tbl:dice_loss}, our experiments with Dice Loss (DL) and Self-Adjusting Dice Loss (SDL) fails to improve above the baseline Cross-Entropy Loss. We test as well as different class-weighting schemes for all of these losses, however, these also do not show improvement. The top-scoring loss was the Vanilla DL formulation, with a generalization scheme proposed in \cite{sudre2017generalised}, however, all trials using DL and DL${(2)}$ are comparable in F1-Score. The self-adjusting Dice Loss, however, underperforms. 

The change made in ADL over DL is the addition of the term $(1 - p_{i, 1})$, which has the effect of downweighting tags that the model is more confident about. This idea has a similar aim as Training Signal Annealing \cite{xie2019unsupervised}, which simply excluding high-confidence predictions. However, ADL has a contradiction: as the model is trained further, it should become more confident; however, as the model becomes more confident, it's confidence is downweighted. It's possible that, with a TSA-like schedule, ADL would not be underperforming as much.

\subsection{Multitask Head Freezing}

Additionally, we experiment with freezing auxiliary heads (heads for tasks that are not \nd) in order to propagate more of the gradient into the shared layers. Note, according to Figure \ref{fig:nn_model}, that this is only the FF layer, which is not a major architectural change. We find that this yields no improvement.

%
%

\end{document}